\documentclass{article}

\usepackage{microtype}
\usepackage{graphicx}
\usepackage{subfigure}
\usepackage{booktabs} %

\usepackage{hyperref}

\usepackage[accepted]{icml2024}

\usepackage{amsmath}
\usepackage{amssymb}
\usepackage{mathtools}
\usepackage{amsthm}

\usepackage{subfiles} %
\usepackage{makecell} %
\usepackage{multirow} %
\usepackage{subcaption}
\usepackage{color,xcolor}
\usepackage{graphicx}
\usepackage{booktabs}
\usepackage{makecell}
\usepackage{colortbl}
\usepackage{pifont}%
\usepackage{algorithm} %
\usepackage{algorithmic} %
\usepackage[switch]{lineno}
\usepackage{amssymb} %

\usepackage[misc]{ifsym}

\usepackage[capitalize,noabbrev]{cleveref}

\theoremstyle{plain}

\theoremstyle{definition}

\theoremstyle{remark}

\usepackage[textsize=tiny]{todonotes}

\definecolor{shapecolor}{rgb}{0.0,0.5,0.0}
\newcommand{\name}{Vim}%

\newcommand{\boldparagraph}[1]{\vspace{0.2cm}\noindent{\bf #1}}

\definecolor{arylideyellow}{rgb}{0.91, 0.84, 0.42}

\def\eg{\emph{e.g.}} 
\def\ie{\emph{i.e.}} 
\def\vs{\emph{vs.}}
\def\etc{\emph{etc.}}

\newcommand{\rblue}{\rowcolor{blue!10}}
\icmltitlerunning{Vision Mamba: Efficient Visual Representation Learning with  
State Space Model}

\begin{document}

\twocolumn[
\icmltitle{Vision Mamba: Efficient Visual Representation Learning with Bidirectional
State Space Model}

\icmlsetsymbol{equal}{*}

\begin{icmlauthorlist}
\icmlauthor{Lianghui Zhu}{eic,equal}
\icmlauthor{Bencheng Liao}{aia,eic,equal}
\icmlauthor{Qian Zhang}{horizon}
\icmlauthor{Xinlong Wang}{baai}
\icmlauthor{Wenyu Liu}{eic}
\icmlauthor{Xinggang Wang}{eic}
\end{icmlauthorlist}

\icmlaffiliation{eic}{School of EIC, Huazhong University of Science \& Technology }
\icmlaffiliation{aia}{Institute of Artificial Intelligence, Huazhong University of Science \& Technology}
\icmlaffiliation{horizon}{Horizon Robotics}
\icmlaffiliation{baai}{Beijing Academy of Artificial Intelligence}

\icmlcorrespondingauthor{Xinggang Wang}{xgwang@hust.edu.cn}

\vskip 0.3in
{%

\vspace{-0.5cm}
\begin{center}
    \centering
    \captionsetup{type=figure}
    \includegraphics[width=1.0\textwidth]{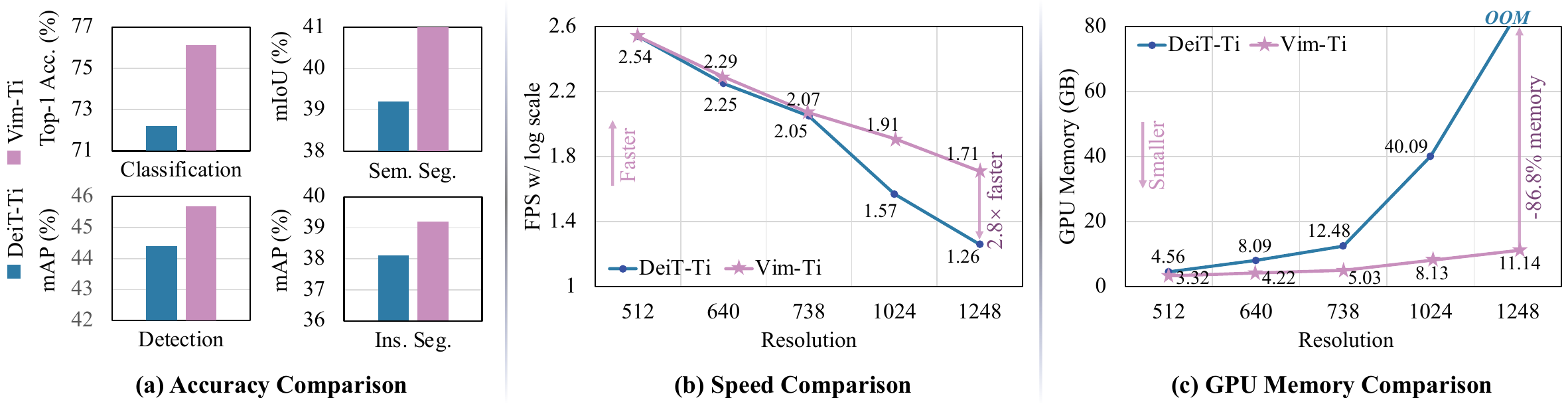}
    \vspace{-.2 in}
    \captionof{figure}{
    Performance and efficiency comparisons between DeiT~\cite{touvron2021deit} and our \name{} model. %
    Results show that \name{} outperforms DeiT on both ImageNet classification and downstream detection and segmentation tasks and is more computation and memory efficient than DeiT in dealing with high-resolution images. For example, \name{} is 2.8$\times$ faster than DeiT and saves 86.8\% GPU memory when performing batch inference to extract features on images with a resolution of 1248$\times$1248, \ie, 6084 tokens per image. 
    }
    \label{fig:vim_teaser}
\end{center}%
}

]

\printAffiliationsAndNotice{\icmlEqualContribution} %

\begin{abstract}
Recently the state space models (SSMs) with efficient hardware-aware designs, i.e., the Mamba deep learning model, have shown great potential for long sequence modeling. Meanwhile building efficient and generic vision backbones purely
upon SSMs is an appealing direction. However, representing visual data is challenging for SSMs due to the position-sensitivity of visual data and
the requirement of global context for visual understanding. In this paper, we show that the reliance on self-attention for visual representation learning is not necessary and propose a new generic vision backbone with bidirectional Mamba blocks
(Vim), which marks the image sequences with position embeddings and compresses the visual representation with bidirectional state space models. On ImageNet classification, COCO object detection, and ADE20k semantic segmentation tasks,
Vim achieves higher performance compared to well-established vision transformers like DeiT, while also demonstrating significantly improved computation \& memory efficiency. For example, Vim is 2.8$\times$ faster than DeiT and saves 86.8\%
GPU memory when performing batch inference to extract features on images with a resolution of 1248$\times$1248. The results demonstrate that Vim is capable of overcoming the computation \& memory constraints on performing Transformer-style understanding for high-resolution images and it has great potential to be the next-generation backbone for vision foundation models.
Code and models are released at \url{https://github.com/hustvl/Vim}
\end{abstract}
\section{Introduction}
Recent research advancements have led to a surge of interest in the state space model (SSM). 
Originating from the classic Kalman filter model \cite{kalman1960filtering}, modern SSMs excel at capturing long-range dependencies and benefit from parallel training.
Some SSM-based methods, such as the linear state-space layers (LSSL)~\cite{gu2021lssl}, structured state space sequence model (S4)~\cite{gu2021s4}, diagonal state space (DSS)~\cite{gupta2022dss}, and S4D~\cite{gu2022s4d}, are proposed to process sequence data across a wide range of tasks and modalities, particularly on modeling long-range dependencies.
They are efficient in processing long sequences because of convolutional computation and near-linear computation. 
2-D SSM~\cite{baron20232dssm}, SGConvNeXt~\cite{li2022sgconvnext}, and ConvSSM~\cite{smith2023convssm} combine SSM with CNN or Transformer architecture to process 2-D data.
The recent work, Mamba~\cite{gu2023mamba}, incorporates time-varying parameters into the SSM and proposes a hardware-aware algorithm to enable very efficient training and inference. 
The superior scaling performance of Mamba indicates that it is a promising alternative to Transformer in language modeling.
Nevertheless, a generic pure-SSM-based backbone network has not been explored for processing visual data, such as images and videos.

Vision Transformers (ViTs) have achieved great success in visual representation learning, excelling in large-scale self-supervised pre-training and high performance on downstream tasks.
Compared with convolutional neural networks, the core advantage lies in that ViT can provide each image patch with data/patch-dependent global context through self-attention. This differs from convolutional networks that use the same parameters, \ie, the convolutional filters, for all positions. Another advantage is the modality-agnostic modeling by treating an image as a sequence of patches without 2D inductive bias, which makes it the preferred architecture for multimodal applications~\cite{fuyu-8b,li2023blip,liu2023visual}. At the same time, the self-attention mechanism in Transformers poses challenges in terms of speed and memory usage when dealing with long-range visual dependencies, \eg, processing high-resolution images. 

Motivated by the success of Mamba in language modeling, it is appealing that we can also transfer this success from language to vision, \ie, to design a generic and efficient visual backbone with the advanced SSM method. However, there are two challenges for Mamba, \ie, unidirectional modeling and lack of positional awareness. 
To address these challenges, we propose the Vision Mamba (\name{}) model, which incorporates the bidirectional SSMs for data-dependent global visual context modeling and position embeddings for location-aware visual recognition. 
We first split the input image into patches and linearly project them as vectors to \name{}. Image patches are treated as the sequence data in \name{} blocks, which efficiently compresses the visual representation with the proposed bidirectional selective state space. Furthermore, the position embedding in \name{} block provides the awareness for spatial information, which enables \name{} to be more robust in dense prediction tasks. 
In the current stage, we train the \name{} model on the supervised image classification task using the ImageNet dataset and then use the pretrained \name{} as the backbone to perform sequential visual representation learning for downstream dense prediction tasks, \ie, semantic segmentation, object detection, and instance segmentation. Like Transformers, \name{} can be pretrained on large-scale unsupervised visual data for better visual representation. Thanks to the better efficiency of Mamba, the large-scale pretraining of \name{} can be achieved with lower computational cost.

Compared with other SSM-based models for vision tasks, \name{} is a pure-SSM-based method and models images in a sequence manner, which is more promising for a generic and efficient backbone. 
Thanks to the bidirectional compressing modeling with positional awareness, \name{} is the first pure-SSM-based model to handle dense prediction tasks. Compared with the most convincing Transformer-based model, \ie, DeiT~\cite{touvron2021deit}, \name{} achieves superior performance on ImageNet classification. Furthermore, \name{} is more efficient in terms of GPU memory and inference time for high-resolution images. The efficiency in terms of memory and speed empowers \name{} to directly perform sequential visual representation learning without relying on 2D priors (such as the 2D local window in ViTDet~\cite{vitdet}) for high-resolution visual understanding tasks while achieving higher accuracy than DeiT.

Our main contributions can be summarized as follows:
\vspace{-4mm}
\begin{itemize}
    \item We propose Vision Mamba (\name{}), which incorporates bidirectional SSM for data-dependent global visual context modeling and position embeddings for location-aware visual understanding.
    \vspace{-2mm}
    \item Without the need of attention, the proposed \name{} has the same modeling power as ViT while it only has subquadratic-time computation and linear memory complexity. Specifically, \name{} is 2.8$\times$ faster than DeiT and saves 86.8\% GPU memory when performing batch inference to extract features on images at the resolution of 1248$\times$1248.
    \vspace{-2mm}
    \item We conduct extensive experiments on ImageNet classification and dense prediction downstream tasks. The results demonstrate that \name{} achieves superior performance compared to the well-established and highly-optimized plain vision Transformer, \ie, DeiT.
\end{itemize}
\section{Related Work}
\boldparagraph{Architectures for generic vision backbone.}
In the early eras, ConvNet~\cite{lecun1998gradient} serves as the de-facto standard network design for computer vision. Many convolutional neural architectures~\cite{krizhevsky2012imagenet,szegedy2015going,simonyan2014very,he2016deep,tan2019efficientnet,wang2020deep,huang2017densely,xie2017aggregated,tan2021efficientnetv2,radosavovic2020designing} have been proposed as the vision backbone for various visual applications. The pioneering work, Vision Transformer (ViT)~\cite{dosovitskiy2020vit} changes the landscape. It treats an image
as a sequence of flattened 2D patches and directly applies a pure Transformer architecture. The surprising results of ViT on image classification and its scaling ability encourage a lot of follow-up works~\cite{touvron2021training,tolstikhin2021mlp,touvron2022resmlp,fang2022msg}. One line of works focuses on hybrid architecture designs by introducing 2D convolutional priors into ViT~\cite{wu2021cvt,dai2021coatnet,d2021convit,dong2022cswin}. PVT~\cite{wang2021pyramid} proposes a pyramid structure Transformer. Swin Transformer~\cite{liu2021swin} applies self-attention within shift windows. Another line of works focuses on improving traditional 2D ConvNets with more advanced settings~\cite{wang2023internimage,liu2022more}. ConvNeXt~\cite{liu2022convnet} reviews the design space and proposes pure ConvNets, which can be scalable as ViT and its variants. RepLKNet~\cite{ding2022scaling} proposes to scale up the kernel size of existing ConvNets to bring improvements.

Though these dominant follow-up works demonstrate superior performance and better efficiency on ImageNet~\cite{deng2009imagenet} and various downstream tasks~\cite{coco,zhou2019ade20k} by introducing 2D priors, with the surge of large-scale visual pretraining \cite{bao2022beit,fang2023eva,caron2021emerging} and multi-modality applications \cite{radford2021learning,li2022blip,li2023blip, liu2023visual,fuyu-8b,jia2021scaling}, vanilla Transformer-style model strikes back to the center stage of computer vision. The advantages of 
larger modeling capacity, unified multi-modality representation, being friendly to self-supervised learning \etc, make it the preferred architecture. However, the number of visual tokens is limited due to the quadratic complexity of Transformer. There are plenty of works~\cite{choromanski2021rethinking,wang2020linformer,Kitaev2020Reformer,child2019generating,ding2023longnet,qin2023hierarchically,retnet} to address this long-standing and prominent challenge, but few of them focus on visual applications. Recently, LongViT~\cite{wang2023image} built an efficient Transformer architecture for computational pathology applications via dilated attention. The linear computation complexity of LongViT allows it to encode the extremely long visual sequence. 
In this work,
we draw inspiration from Mamba~\cite{gu2023mamba} and explore building a pure-SSM-based model as a generic vision backbone without using attention, while preserving the sequential, modality-agnostic modeling merit of ViT.

\boldparagraph{State space models for long sequence modeling.}
\cite{gu2021s4} proposes a Structured State-Space Sequence (S4) model, a novel alternative to
CNNs or Transformers, to model the long-range dependency. The promising property of linearly scaling in sequence length attracts further explorations.
\cite{wang2022pretraining} proposes Bidirectional Gated SSM to replicate BERT~\cite{devlin2018bert} results without attention.
\cite{smith2023simplified} proposes a new S5 layer by introducing MIMO SSM and efficient parallel scan into S4 layer. 
\cite{fu2023hungry} designs a new SSM layer, H3, that nearly fills the performance gap between SSMs and Transformer attention in language modeling. 
\cite{mehta2023long} builds the Gated State Space layer on S4 by introducing more gating units to improve the expressivity.
Recently, \cite{gu2023mamba} proposes a data-dependent SSM layer and builds a generic language model backbone, Mamba, which outperforms Transformers at various sizes on large-scale real data and enjoys linear scaling in sequence length. In this work, we explore transferring the success of Mamba to vision, \ie, building a generic vision backbone purely upon SSM without attention.

\boldparagraph{State space models for visual applications.}
\cite{islam2022long} uses 1D S4 to handle the long-range temporal dependencies for video classification. \cite{nguyen2022s4nd} further extends 1D S4 to handle multi-dimensional data including 2D images and 3D videos.
\cite{islam2023efficient} combines the strengths of S4 and self-attention to build TranS4mer model, achieving state-of-the-art performance for movie scene detection. 
\cite{wang2023selective} introduces a novel selectivity mechanism to S4, largely improving the performance of S4 on long-form video understanding with a much lower memory footprint.
\cite{yan2023diffusion} supplants attention mechanisms with a more scalable SSM-based backbone to generate high-resolution images and process fine-grained representation under affordable computation.
\cite{ma2024u} proposes U-Mamba, a hybrid CNN-SSM architecture, to handle the long-range dependencies in biomedical image segmentation. The above works~\cite{xing2024segmamba,ma2024u,yan2023diffusion,wang2023selective,islam2023efficient,nguyen2022s4nd,islam2022long} either apply SSM to specific visual applications or build a hybrid architecture by combining SSM with convolution or attention. Different from them, we build a pure-SSM-based model, which can be adopted as a generic vision backbone. It is noteworthy that VMamba \cite{vmamba}, a concurrent work with our method, has demonstrated impressive results in visual recognition by incorporating Mamba with multi-directional scanning and a hierarchical network architecture. In contrast, \name{} primarily concentrates on visual sequence learning and boasts a unified representation for multi-modality data.
\section{Method}
The goal of Vision Mamba (\name{}) is to introduce the advanced state space model (SSM), \ie, Mamba~\cite{gu2023mamba}, to computer vision. This section begins with a description of the preliminaries of SSM. It is followed by an overview of \name{}. We then detail how the \name{} block processes input token sequences and proceed to illustrate the architecture details of \name{}. The section concludes with an analysis of the efficiency of the proposed \name{}.

\begin{figure*}[t]
    \centering
    \includegraphics[width=1.0\linewidth]{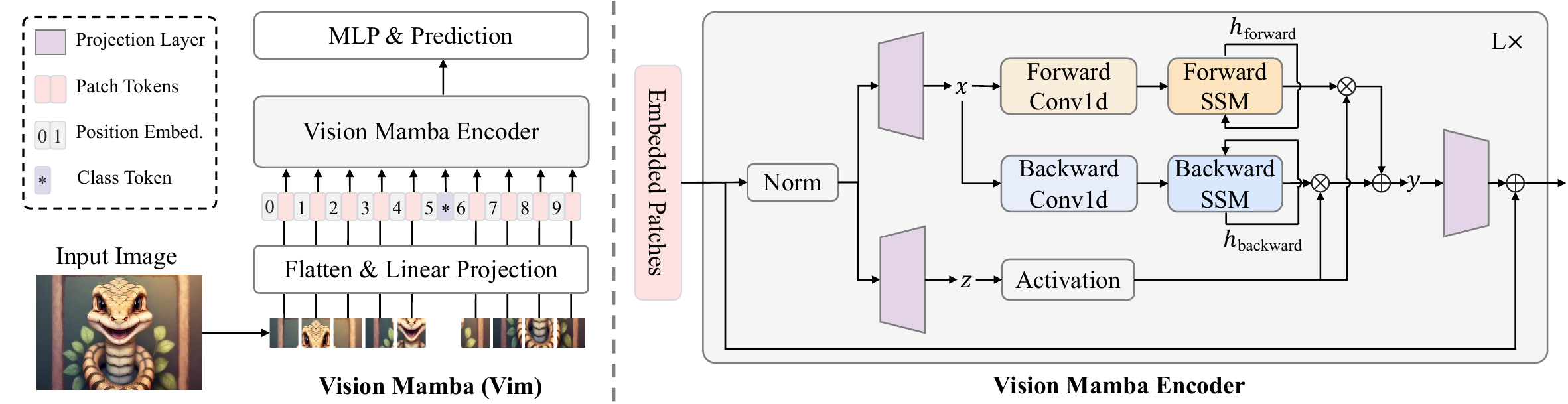}
    \caption{The overview of the proposed \name{} model. We first split the input image into patches, and then project them into patch tokens. Last, we send the sequence of tokens to the proposed \name{} encoder. To perform ImageNet classification, we concatenate an extra learnable classification token to the patch token sequence. 
    Different from Mamba for text sequence modeling, \name{} encoder processes the token sequence with both forward and backward directions.
    }
    \label{fig:pipeline}
\vspace{-0.1 in}
\end{figure*}

\subsection{Preliminaries}
\label{sec:pre}

The SSM-based models, \ie, structured state space sequence models (S4) and Mamba are inspired by the continuous system, which maps a 1-D function or sequence $x(t) \in \mathbb{R} \mapsto y(t) \in \mathbb{R}$ through a hidden state $h(t) \in \mathbb{R}^\mathtt{N}$. This system uses $\mathbf{A} \in \mathbb{R}^{\mathtt{N} \times \mathtt{N}}$ as the evolution parameter and $\mathbf{B} \in \mathbb{R}^{\mathtt{N} \times 1}$, $\mathbf{C} \in \mathbb{R}^{1 \times \mathtt{N}}$ as the projection parameters. The continuous system works as follows: $h'(t) = \mathbf{A}h(t) + \mathbf{B}x(t)$ and $y(t) = \mathbf{C}h(t)$.

The S4 and Mamba are the discrete versions of the continuous system, which include a timescale parameter $\mathbf{\Delta}$ to transform the continuous parameters $\mathbf{A}$, $\mathbf{B}$ to discrete parameters $\mathbf{\overline{A}}$, $\mathbf{\overline{B}}$. The commonly used method for transformation is zero-order hold (ZOH), which is defined as follows:
\begin{equation}
\begin{aligned}
\label{eq:zoh}
\mathbf{\overline{A}} &= \exp{(\mathbf{\Delta}\mathbf{A})}, \\
\mathbf{\overline{B}} &= (\mathbf{\Delta} \mathbf{A})^{-1}(\exp{(\mathbf{\Delta} \mathbf{A})} - \mathbf{I}) \cdot \mathbf{\Delta} \mathbf{B}.
\end{aligned}
\end{equation}
After the discretization of $\mathbf{\overline{A}}$, $\mathbf{\overline{B}}$, the discretized version  using a step size $\mathbf{\Delta}$ can be rewritten as:
\begin{equation}
\begin{aligned}
\label{eq:discrete_lti}
h_t &= \mathbf{\overline{A}}h_{t-1} + \mathbf{\overline{B}}x_{t}, \\
y_t &= \mathbf{C}h_t.
\end{aligned}
\end{equation}
At last, the models compute output through a global convolution. 
\begin{equation}
\begin{aligned}
\label{eq:conv}
\mathbf{\overline{K}} &= (\mathbf{C}\mathbf{\overline{B}}, \mathbf{C}\mathbf{\overline{A}}\mathbf{\overline{B}}, \dots, \mathbf{C}\mathbf{\overline{A}}^{\mathtt{M}-1}\mathbf{\overline{B}}), \\
\mathbf{y} &= \mathbf{x} * \mathbf{\overline{K}},
\end{aligned}
\end{equation}
where $\mathtt{M}$ is the length of the input sequence $\mathbf{x}$, and $\overline{\mathbf{K}} \in \mathbb{R}^{\mathtt{M}}$ is a structured convolutional kernel.

\subsection{Vision Mamba}
\label{sec:vim}
An overview of the proposed \name{} is shown in Fig.~\ref{fig:pipeline}. The standard Mamba is designed for the 1-D sequence. To process the vision tasks, we first transform the 2-D image $\mathbf{t} \in \mathbb{R}^{\mathtt{H} \times \mathtt{W} \times \mathtt{C}}$ into the flattened 2-D patches $\mathbf{x_p} \in \mathbb{R}^{\mathtt{J} \times (\mathtt{P}^2 \cdot  \mathtt{C})}$, where $(\mathtt{H}, \mathtt{W})$ is the size of input image, $\mathtt{C}$ is the number of channels, $\mathtt{P}$ is the size of image patches. Next, we linearly project the $\mathbf{x_p}$ to the vector with size $\mathtt{D}$ and add position embeddings $\mathbf{E}_{pos} \in \mathbb{R}^{(\mathtt{J}+1) \times \mathtt{D}}$, as follows:
\begin{equation}
\begin{aligned}
\label{eq:embed}
\mathbf{T}_0 &= [\mathbf{t}_{cls};\mathbf{t}_p^1\mathbf{W};\mathbf{t}_p^2\mathbf{W};\cdots;\mathbf{t}_p^{\mathtt{J}}\mathbf{W}] + \mathbf{E}_{pos}, \\
\end{aligned}
\end{equation}
where $\mathbf{t}_p^{\mathtt{j}}$ is the $\mathtt{j}$-th patch of $\mathbf{t}$, $\mathbf{W} \in \mathbb{R}^{(\mathtt{P}^2 \cdot \mathtt{C}) \times \mathtt{D}}$ is the learnable projection matrix. Inspired by ViT~\cite{dosovitskiy2020vit} and BERT~\cite{kenton2019bert}, we also use class token to represent the whole patch sequence, which is denoted as $\mathbf{t}_{cls}$. We then send the token sequence ($\mathbf{T}_{\mathtt{l}-1}$) to the $\mathtt{l}$-th layer of the \name{} encoder, and get the output $\mathbf{T}_{\mathtt{l}}$. Finally, we normalize the output class token $\mathbf{T}_{\mathtt{L}}^0$ and feed it to the multi-layer perceptron (MLP) head to get the final prediction $\hat{p}$, as follows: $\mathbf{T}_l = \mathbf{\name{}}(\mathbf{T}_{\mathtt{l}-1}) + \mathbf{T}_{\mathtt{l}-1}$, $\mathbf{f} = \mathbf{Norm}(\mathbf{T}_{\mathtt{L}}^0)$, and $\hat{p} = \mathbf{MLP}(\mathbf{f})$, where $\mathbf{\name{}}$ is the proposed vision mamba block, $\mathtt{L}$ is the number of layers, and $\mathbf{Norm}$ is the normalization layer.

\subsection{\name{} Block} 
\label{sec:vim_block}

The original Mamba block is designed for the 1-D sequence, which is not suitable for vision tasks requiring spatial-aware understanding. In this section, we introduce the \name{} block, which incorporates the bidirectional sequence modeling for the vision tasks. The \name{} block is shown in Fig.~\ref{fig:pipeline}.

Specifically, we present the operations of \name{} block in Algo.~\ref{alg:block}. The input token sequence $\mathbf{T}_{\mathtt{l}-1}$ is first normalized by the normalization layer. Next, we linearly project the normalized sequence to the $\mathbf{x}$ and $\mathbf{z}$ with dimension size $E$. Then, we process the $\mathbf{x}$ from the forward and backward directions. 
For each direction, we first apply the 1-D convolution to the $\mathbf{x}$ and get the $\mathbf{x}'_{o}$. We then linearly project the $\mathbf{x}'_{o}$ to the $\mathbf{B}_{o}$, $\mathbf{C}_{o}$, $\mathbf{\Delta}_{o}$, respectively. 
The $\mathbf{\Delta}_{o}$ is then used to transform the $\overline{\mathbf{A}}_{o}$, $\overline{\mathbf{B}}_{o}$, respectively. 
Finally, we compute the $\mathbf{y}_{forward}$ and $\mathbf{y}_{backward}$ through the SSM. The $\mathbf{y}_{forward}$ and $\mathbf{y}_{backward}$ are then gated by the $\mathbf{z}$ and added together to get the output token sequence $\mathbf{T}_{\mathtt{l}}$.

\subsection{Architecture Details}

In summary, the hyper-parameters of our architecture are listed as follows: $\mathtt{L}$ denotes the number of blocks, $\mathtt{D}$ denotes the hidden state dimension, $\mathtt{E}$ denotes the expanded state dimension, and $\mathtt{N}$ denotes the SSM dimension.
Following ViT~\cite{dosovitskiy2020vit} and DeiT~\cite{touvron2021training}, we first employ 16$\times$16 kernel size projection layer to get a 1-D sequence of non-overlapping patch embeddings. Subsequently, we directly stack  $\mathtt{L}$ Vim blocks. By default, we set the number of blocks $\mathtt{L}$ to 24, SSM dimension $\mathtt{N}$ to 16. To align with the model sizes of DeiT series, we set the hidden state dimension $\mathtt{D}$ to 192 and expanded state dimension $\mathtt{E}$ to 384 for the tiny-size variant. For the small-size variant, we set $\mathtt{D}$ to 384 and $\mathtt{E}$ to 768.

\begin{algorithm}[h]
\caption{\name{} Block Process}
\small
\begin{algorithmic}[1]
\REQUIRE{token sequence $\mathbf{T}_{l-1}$ : \textcolor{shapecolor}{$(\mathtt{B}, \mathtt{M}, \mathtt{D})$}}
\ENSURE{token sequence $\mathbf{T}_{l}$ : \textcolor{shapecolor}{$(\mathtt{B}, \mathtt{M}, \mathtt{D})$}}
\STATE \textcolor{gray}{\text{/* normalize the input sequence $\mathbf{T}_{l-1}'$ */}}
\STATE $\mathbf{T}_{l-1}'$ : \textcolor{shapecolor}{$(\mathtt{B}, \mathtt{M}, \mathtt{D})$} $\leftarrow$ $\mathbf{Norm}(\mathbf{T}_{l-1})$
\STATE $\mathbf{x}$ : \textcolor{shapecolor}{$(\mathtt{B}, \mathtt{M}, \mathtt{E})$} $\leftarrow$ $\mathbf{Linear}^\mathbf{x}(\mathbf{T}_{l-1}')$
\STATE $\mathbf{z}$ : \textcolor{shapecolor}{$(\mathtt{B}, \mathtt{M}, \mathtt{E})$} $\leftarrow$ $\mathbf{Linear}^\mathbf{z}(\mathbf{T}_{l-1}')$
\STATE \textcolor{gray}{\text{/* process with different direction */}}
\FOR{$o$ in \{forward, backward\}}
\STATE $\mathbf{x}'_o$ : \textcolor{shapecolor}{$(\mathtt{B}, \mathtt{M}, \mathtt{E})$} $\leftarrow$ $\mathbf{SiLU}(\mathbf{Conv1d}_o(\mathbf{x}))$
\STATE $\mathbf{B}_o$ : \textcolor{shapecolor}{$(\mathtt{B}, \mathtt{M}, \mathtt{N})$} $\leftarrow$ $\mathbf{Linear}_o^{\mathbf{B}}(\mathbf{x}'_o)$
\STATE $\mathbf{C}_o$ : \textcolor{shapecolor}{$(\mathtt{B}, \mathtt{M}, \mathtt{N})$} $\leftarrow$ $\mathbf{Linear}^{\mathbf{C}}_o(\mathbf{x}'_o)$
\STATE \textcolor{gray}{\text{/* softplus ensures positive $\mathbf{\Delta}_o$ */}}
\STATE $\mathbf{\Delta}_o$ : \textcolor{shapecolor}{$(\mathtt{B}, \mathtt{M}, \mathtt{E})$} $\leftarrow$ $\log(1 + \exp(\mathbf{Linear}_o^{\mathbf{\Delta}}(\mathbf{x}'_o) + \mathbf{Parameter}_o^{\mathbf{\Delta}}))$
\STATE \textcolor{gray}{\text{/* shape of $\mathbf{Parameter}_o^{\mathbf{A}}$ is \textcolor{shapecolor}{$(\mathtt{E}, \mathtt{N})$} */}}
\STATE $\overline{\mathbf{A}_o}$ : \textcolor{shapecolor}{$(\mathtt{B}, \mathtt{M}, \mathtt{E}, \mathtt{N})$} $\leftarrow$ $\mathbf{\Delta}_o \bigotimes \mathbf{Parameter}_o^{\mathbf{A}}$ 
\STATE $\overline{\mathbf{B}_o}$ : \textcolor{shapecolor}{$(\mathtt{B}, \mathtt{M}, \mathtt{E}, \mathtt{N})$} $\leftarrow$ $\mathbf{\Delta}_o \bigotimes \mathbf{B}_o$
\STATE \textcolor{gray}{\text{/* initialize $h_o$ and $\mathbf{y}_o$ with $0$ */}}
\STATE $h_o$ : \textcolor{shapecolor}{$(\mathtt{B}, \mathtt{E}, \mathtt{N})$} $\leftarrow$ zeros \textcolor{shapecolor}{$(\mathtt{B}, \mathtt{E}, \mathtt{N})$}
\STATE $\mathbf{y}_o$ : \textcolor{shapecolor}{$(\mathtt{B}, \mathtt{M}, \mathtt{E})$} $\leftarrow$ zeros \textcolor{shapecolor}{$(\mathtt{B}, \mathtt{M}, \mathtt{E})$}
\STATE \textcolor{gray}{\text{/* SSM recurrent */}}
\FOR{$i$ in \{0, ..., M-1\}}
\STATE $h_o$ = $\overline{\mathbf{A}_o}[:,i,:,:] \bigodot h_o + \overline{\mathbf{B}_o}[:,i,:,:] \bigodot \mathbf{x}_o'[:,i,:,\textcolor{shapecolor}{\mathtt{None}}] $
\STATE $\mathbf{y}_o[:,i,:]$ = $h_o \bigotimes \mathbf{C}_o[:,i,:]$
\ENDFOR
\ENDFOR
\STATE \textcolor{gray}{\text{/* get gated $\mathbf{y}$ */}}
\STATE $\mathbf{y}_{forward}'$ : \textcolor{shapecolor}{$(\mathtt{B}, \mathtt{M}, \mathtt{E})$} $\leftarrow$ $\mathbf{y}_{forward} \bigodot \mathbf{SiLU}(\mathbf{z}) $
\STATE $\mathbf{y}_{backward}'$ : \textcolor{shapecolor}{$(\mathtt{B}, \mathtt{M}, \mathtt{E})$} $\leftarrow$ $\mathbf{y}_{backward} \bigodot \mathbf{SiLU}(\mathbf{z}) $
\STATE \textcolor{gray}{\text{/* residual connection */}}
\STATE $\mathbf{T}_{l}$ : \textcolor{shapecolor}{$(\mathtt{B}, \mathtt{M}, \mathtt{D})$} $\leftarrow$ $\mathbf{Linear}^\mathbf{T}(\mathbf{y}_{forward}' + \mathbf{y}_{backward}') + \mathbf{T}_{l-1}$
\STATE Return: $\mathbf{T}_{l}$ 
\label{alg:block}
\end{algorithmic}
\end{algorithm}

\subsection{Efficiency Analysis}
\label{sec:efficiency}
Traditional SSM-based methods leverage the fast Fourier transform to boost the convolution operation as shown in Eq.~\eqref{eq:conv}. For data-dependent methods, such as Mamba, the SSM operation in Line 11 of Algo.~\ref{alg:block} is no longer equivalent to convolution. To address this problem, Mamba and the proposed \name{} choose a modern-hardware-friendly way to ensure efficiency. The key idea of this optimization is to avoid the IO-bound and memory-bound of modern hardware accelerators (GPUs).

\boldparagraph{IO-Efficiency.}
The high bandwidth memory (HBM) and SRAM are two important components for GPUs. Among them, SRAM has a larger bandwidth and HBM has a bigger memory size. The standard implementation of \name's SSM operation with HBM requires the number of memory IO on the order of $O(\mathtt{B}\mathtt{M}\mathtt{E}\mathtt{N})$. Inspired by Mamba, \name{} first reads in $O(\mathtt{B}\mathtt{M}\mathtt{E} + \mathtt{E}\mathtt{N})$ bytes of memory $(\mathbf{\Delta_o}, \mathbf{{A}_o}, \mathbf{{B}_o}, \mathbf{C_o)}$ from slow HBM to fast SRAM. Then, \name{} gets the discrete $\mathbf{\overline{A}_o}$, $\mathbf{\overline{B}_o}$ of a size of $(\mathtt{B}, \mathtt{M}, \mathtt{E}, \mathtt{N})$ in SRAM. Last, \name{} performs SSM operations in SRAM and writes the output of a size of $(\mathtt{B}, \mathtt{M}, \mathtt{E})$ back to HBM. This method can help to reduce IOs from $O(\mathtt{B}\mathtt{M}\mathtt{E}\mathtt{N})$ to $O(\mathtt{B}\mathtt{M}\mathtt{E} + \mathtt{E}\mathtt{N})$.

\boldparagraph{Memory-Efficiency.}
To avoid out-of-memory problems and achieve lower memory usage when dealing with long sequences, \name{} chooses the same recomputation method as Mamba. For the intermediate states of size $(\mathtt{B}, \mathtt{M}, \mathtt{E}, \mathtt{N})$ to calculate the gradient, \name{} recomputes them at the network backward pass. For intermediate activations such as the output of activation functions and convolution, \name{} also recomputes them to optimize the GPU memory requirement, as the activation values take a lot of memory but are fast for recomputation.

\boldparagraph{Computation-Efficiency.}
SSM in Vim block (Line 11 in Algo.\ref{alg:block}) and self-attention in Transformer both play a key role in providing global context adaptively. Given a visual sequence $\mathbf{T} \in R^{1\times \mathtt{M} \times \mathtt{D}} $ and the default setting $\mathtt{E}=2\mathtt{D}$, the computation complexity of a global self-attention and SSM are:
\begin{align}
\label{eq:self-attn}
&\Omega (\text{self-attention}) = 4\mathtt{M}\mathtt{D}^2 + 2\mathtt{M}^2\mathtt{D}, \\
&\Omega (\text{SSM}) = 3\mathtt{M}(2\mathtt{D})\mathtt{N} + \mathtt{M}(2\mathtt{D})\mathtt{N}, 
\end{align}
where self-attention is quadratic to sequence length $\mathtt{M}$, and SSM is linear to sequence length $\mathtt{M}$ ($\mathtt{N}$ is a fixed parameter, set to 16 by default). The computational efficiency makes \name{} scalable for gigapixel applications with large sequence lengths.

\section{Experiment}
\begin{table}[h]
\centering
\small
\begin{tabular}{l | c  c  | c }
\toprule
Method &  \begin{tabular}[c]{@{}c@{}}image \\ size\end{tabular} & \#param.  & \begin{tabular}[c]{@{}c@{}}ImageNet \\ top-1 acc.\end{tabular} \\
\toprule
\multicolumn{4}{c}{\textbf{Convnets}} \\
\midrule
ResNet-18 & $224^{2}$ & 12M  & 69.8\\
ResNet-50 & $224^{2}$ & 25M  & 76.2\\
ResNet-101 & $224^{2}$& 45M  & 77.4\\
ResNet-152 &$224^{2}$ & 60M  & 78.3\\
\midrule
ResNeXt50-32$\times$4d & $224^{2}$ & 25M & 77.6 \\
\midrule
RegNetY-4GF & $224^{2}$ & 21M &80.0 \\

\toprule
\multicolumn{4}{c}{\textbf{Transformers}} \\
\midrule
ViT-B/16 &$384^{2}$ &  86M & 77.9  \\
ViT-L/16 & $384^{2}$ & 307M  & 76.5 \\
\midrule
DeiT-Ti & $224^{2}$ & 6M  & 72.2  \\
DeiT-S & $224^{2}$ & 22M  & 79.8 \\
DeiT-B &  $224^{2}$ & 86M  & 81.8 \\
\toprule
\multicolumn{4}{c}{\textbf{SSMs}} \\
\midrule
S4ND-ViT-B & $224^{2}$ & 89M & 80.4 \\
\midrule
\rblue
\name{}-Ti &$224^{2}$ & 7M & 76.1 \\ %
\rblue
\name{}-Ti$^\dagger$ & $224^{2}$ & 7M & 78.3 \scriptsize\color{red}{+2.2}\\ %
\midrule
\rblue
\name{}-S & $224^{2}$ & 26M & 80.3 \\
\rblue
\name{}-S$^\dagger$ & $224^{2}$& 26M & 81.4 \scriptsize\color{red}{+1.1}\\ %
\midrule
\rblue
\name{}-B & $224^{2}$ & 98M & 81.9 \\
\rblue
\name{}-B$^\dagger$ & $224^{2}$ & 98M & 83.2 \scriptsize\color{red}{+1.3}\\
\bottomrule
\end{tabular}
\caption{Comparison with different backbones on ImageNet-1K validation set. $^\dagger$ represents the model is fine-tuned with our long sequence setting.}
\label{tab:clscomp}
\vspace{-0.1 in}
\end{table}

\subsection{Image Classification}
\vspace{-.1 in}
\boldparagraph{Settings.} We benchmark \name{} on the ImageNet-1K dataset~\cite{deng2009imagenet}, which contains 1.28M training images and 50K validation images from 1,000 categories. All models are trained on the training set, and top-1 accuracy on the validation set is reported. For fair comparisons, our training settings mainly follow DeiT~\cite{touvron2021training}. Specifically, we apply random cropping, random horizontal flipping, label-smoothing regularization, mixup, and random erasing as data augmentations. When training on $224^2$ input images, we employ AdamW~\cite{adamw} with a momentum of $0.9$, a total batch size of $1024$, and a weight decay of $0.05$ to optimize models. We train the \name{} models for $300$ epochs using a cosine schedule, $1\times$$10^{-3}$ initial learning rate, and EMA. During testing, we apply a center crop on the validation set to crop out $224^2$ images. Experiments are performed on 8 A800 GPUs.

\begin{table}[htp]
\centering
\small
\begin{tabular}{l c | c  c  | c }
\toprule
Method &  Backbone & \begin{tabular}[c]{@{}c@{}}image \\ size\end{tabular} & \#param.  & \begin{tabular}[c]{@{}c@{}}$val$ \\ mIoU\end{tabular} \\
\toprule
DeepLab v3+ & ResNet-101&$512^{2}$ & 63M  & 44.1\\ %
UperNet  & ResNet-50 &$512^{2}$ & 67M  & 41.2 \\ %
UperNet  & ResNet-101 &$512^{2}$ & 86M  & 44.9 \\ %
\midrule
UperNet  & DeiT-Ti&$512^{2}$ & 11M  & 39.2  \\ %
UperNet  & DeiT-S&$512^{2}$ & 43M  & 44.0 \\ %
\midrule
\rblue
UperNet  & \name{}-Ti &$512^{2}$ & 13M & 41.0 \\%12520940
\rblue
UperNet  & \name{}-S &$512^{2}$ & 46M & 44.9 \\%45867692
\bottomrule
\end{tabular}
\caption{Results of semantic segmentation on the ADE20K $val$ set. }
\label{tab:segcomp}
\vspace{-0.35 in}
\end{table}

\boldparagraph{Long Sequence Fine-tuning} To make full use of the efficient long sequence modeling power of \name{}, we continue to fine-tune \name{} with a long sequence setting for 30 epochs after pretraining. Specifically, we set a patch extraction stride of $8$ while keeping the patch size unchanged, a constant learning rate of $10^{-5}$, and a weight decay of $10^{-8}$. 

\boldparagraph{Results.} Tab.~\ref{tab:clscomp} compares \name{} with ConvNet-based, Transformer-based and SSM-based backbone networks. Compared to ConvNet-based ResNet~\cite{he2016deep}, \name{} demonstrates superior performance. For example, when the parameters are roughly similar, the top-1 accuracy of \name{}-Small reaches 80.3, which is 4.1 points higher than that of ResNet50. Compared with the conventional self-attention-based ViT~\cite{dosovitskiy2020vit}, \name{} outperforms it by considerable margins in terms of both parameter numbers and classification accuracy. When compared to the highly-optimized ViT-variant, \ie, DeiT~\cite{touvron2021training}, \name{} surpasses it at different scales with comparable parameter numbers: 3.9 points higher for \name{}-Tiny over DeiT-Tiny, 0.5 points higher for \name{}-Small over DeiT-Small, and 0.1 points higher for \name{}-Base over DeiT-Base. 
Compared with SSM-based S4ND-ViT-B~\cite{nguyen2022s4nd}, \name{} achieves similar top-1 accuracy with 3$\times$ fewer parameters. After long sequence fine-tuning, \name{}-Tiny$^\dagger$, \name{}-S$^\dagger$, and \name{}-B$^\dagger$ all achieve higher results. Among them, \name{}-S$^\dagger$ even achieves similar results with DeiT-B. 
The results demonstrate that \name{} can be adapted to longer sequence modeling easily and extract stronger visual representation.

Fig.~\ref{fig:vim_teaser} (b) and (c) compare the FPS and GPU memory of tiny-size \name{} and DeiT. \name{} demonstrates better efficiency in speed and memory as image resolution grows. Specifically,  when the image size is 512$\times$512, \name{} achieves similar FPS and memory as DeiT. As the image size grows to 1248$\times$1248, \name{} is 2.8$\times$ faster than DeiT and saves 86.8\% GPU memory. The pronounced superiority of Vim's linear scaling in sequence length makes it ready for high-resolution downstream vision applications and long-sequence multi-modality applications.

\begin{table}
\centering
\small
\addtolength{\tabcolsep}{-1pt}
\begin{tabular}{m{40pt} | m{20pt}m{20pt}m{22pt} | m{20pt}m{20pt}m{22pt}}
\toprule
Backbone & AP$^{\text{box}}$ & AP$^{\text{box}}_{\text{50}}$  & AP$^{\text{box}}_{\text{75}}$ & AP$^{\text{box}}_{\text{s}}$ & AP$^{\text{box}}_{\text{m}}$  & AP$^{\text{box}}_{\text{l}}$ \\
\midrule
DeiT-Ti & 44.4  & 63.0 & 47.8 & 26.1  & 47.4 & 61.8   \\
\rblue
\name{}-Ti & 45.7 & 63.9 & 49.6 & 26.1 & 49.0 & 63.2\\
\toprule
Backbone & AP$^{\text{mask}}$ & AP$^{\text{mask}}_{\text{50}}$  & AP$^{\text{mask}}_{\text{75}}$ & AP$^{\text{mask}}_{\text{s}}$ & AP$^{\text{mask}}_{\text{m}}$  & AP$^{\text{mask}}_{\text{l}}$ \\
\midrule
DeiT-Ti & 38.1 & 59.9 & 40.5 & 18.1  & 40.5 & 58.4   \\
\rblue
\name{}-Ti &39.2 &60.9& 41.7& 18.2 & 41.8 & 60.2 \\
\bottomrule
\end{tabular}
\caption{Results of object detection and instance segmentation on the COCO $val$ set using Cascade Mask R-CNN~\cite{cai2019cmrcnn} framework. }
\label{tab:detcomp}
\vspace{-0.2 in}
\end{table}

\begin{figure}[t]
    \centering
    \includegraphics[width=1.\linewidth]{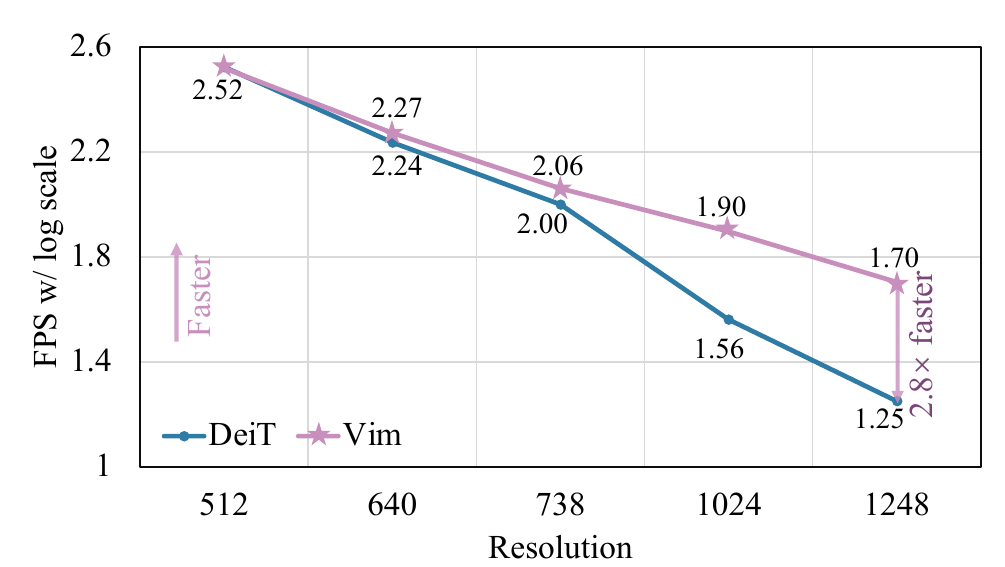}
    \caption{FPS comparison between DeiT-Ti~\cite{touvron2021deit} and our \name{}-Ti on the commonly used downstream framework. We perform batch inference and benchmark the log-scaled FPS on the architecture with the backbone and FPN. \name{} achieves comparable performance to DeiT with a small resolution, \ie, 512$\times$512. As the input image resolution increases, \name{} has a higher FPS.
    }
    \label{fig:fpscomp}
\end{figure}

\subsection{Semantic Segmentation}

\boldparagraph{Settings.} We conduct experiments for semantic segmentation on the ADE20K~\cite{zhou2019ade20k} and use UperNet~\cite{xiao2018upernet} as the segmentation framework. We provide detailed settings in Sec.~\ref{sec: add_setting}.
\boldparagraph{Results.} As shown in Tab.~\ref{tab:segcomp}, \name{} consistently outperforms DeiT across different scales: 1.8 mIoU higher for \name{}-Ti over DeiT-Ti, and 0.9 mIoU higher for \name{}-S over DeiT-S. Compared to the ResNet-101 backbone, our \name{}-S achieves the same segmentation performance with nearly 2$\times$ fewer parameters. 

To further evaluate the efficiency for downstream tasks, \ie, segmentation, detection, and instance segmentation, we combine the backbones with a commonly used feature pyramid network (FPN) module and benchmark their FPS and GPU memory.
As shown in Fig.~\ref{fig:fpscomp} and Fig.~\ref{fig:memcomp}, the efficiency curves demonstrate similar comparison results of the pure backbone (Fig.~\ref{fig:vim_teaser}), though we append a heavy FPN on the backbones.
The exceptional linear scaling performance is attributed to our proposed efficient backbone Vim, which builds the foundation for learning gigapixel-level visual representation in an end-to-end manner without the need for multi-stage encoding (\eg, aerial image, medical image, and computational pathology).

\subsection{Object Detection and Instance Segmentation}

\boldparagraph{Settings.} We conduct experiments for object detection and instance segmentation on the COCO 2017 dataset~\cite{coco} and use ViTDet~\cite{xiao2018upernet} as the basic framework. We provide detailed settings in Sec.~\ref{sec: add_setting}.

\begin{figure}[t]
    \centering
    \includegraphics[width=1.\linewidth]{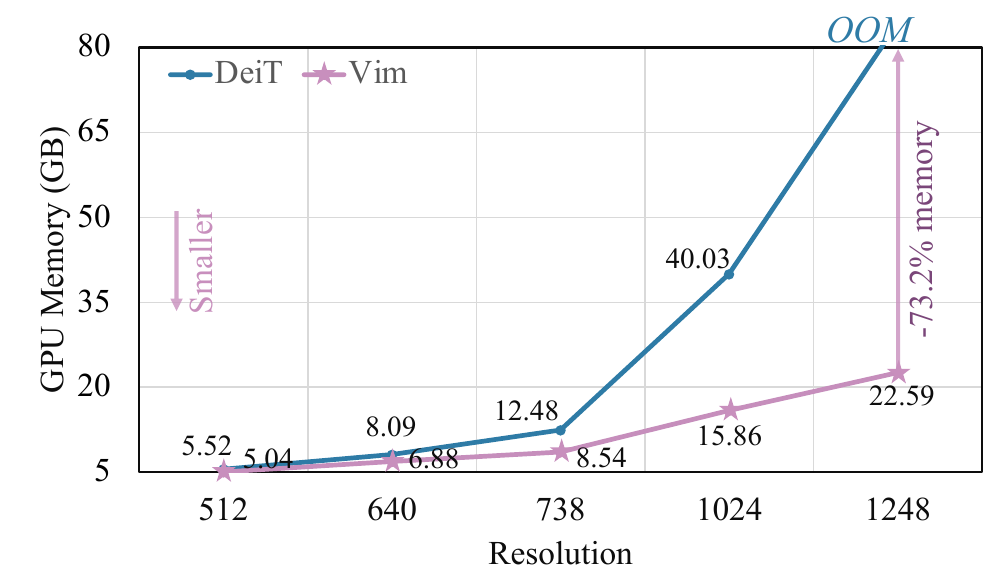}
    \caption{GPU memory efficiency comparison between DeiT-Ti~\cite{touvron2021deit} and our \name{}-Ti on the commonly used downstream framework. We perform batch inference and benchmark the GPU memory on the architecture with the backbone and FPN. \name{} requires comparable GPU memory to DeiT with a small resolution, \ie, 512$\times$512. As the input image resolution increases, \name{} will use significantly less GPU memory.
    }
    \label{fig:memcomp}
\end{figure}

\boldparagraph{Results.} Tab.~\ref{tab:detcomp} compares Vim-Ti with DeiT-Ti using Cascade Mask R-CNN framework~\cite{cai2019cmrcnn}. Vim-Ti surpasses DeiT-Ti by 1.3 box AP and 1.1 mask AP. For the middle-size and large-size objects, Vim-Ti outperforms DeiT-Ti by 1.6 AP$^{\text{box}}_{\text{m}}$/1.3 AP$^{\text{mask}}_{\text{m}}$ and 1.4 AP$^{\text{box}}_{\text{l}}$/1.8 AP$^{\text{mask}}_{\text{l}}$, demonstrating better long-range context learning than DeiT (Fig.~\ref{fig:viscomp}).

We highlight that the accuracy superiority is non-trivial since DeiT is equipped with window attention while \name{} works in a pure sequence modeling manner. Specifically, to perform representation learning on high-resolution images (\ie, 1024$\times$1024), we follow ViTDet~\cite{vitdet} and modify the DeiT backbone with the use of 2D window attention, which injects 2D prior and breaks the sequential modeling nature of Transformer. Thanks to the efficiency illustrated in Sec.~\ref{sec:efficiency}, Fig.~\ref{fig:vim_teaser} and Fig.~\ref{fig:memcomp}, we can directly apply Vim on 1024$\times$1024 input images and learn sequential visual representation for object detection and instance segmentation without need for 2D priors in the backbone. 

\begin{table}
\centering
\small
\begin{tabular}{l  | c  | c }
\toprule
Bidirectional strategy  & \begin{tabular}[c]{@{}c@{}}ImageNet \\ top-1 acc.\end{tabular} & \begin{tabular}[c]{@{}c@{}}ADE20K \\ mIoU\end{tabular} \\
\midrule
\textit{None}& 73.2 & 32.3\\
\midrule
\textit{Bidirectional Layer} & 70.9 & 33.6 \\
\textit{Bidirectional SSM} & 72.8 & 33.2 \\
\rblue
\textit{Bidirectional SSM + Conv1d} & 73.9 & 35.9\\
\bottomrule
\end{tabular}
\caption{Ablation study on the bidirectional design. To ensure a fair comparison, we do not use the class token for each experiment. The default setting for \name{} is marked in \colorbox{blue!10}{blue}.}
\label{tab:aba-dir}
\end{table}

\subsection{Ablation Study}
\boldparagraph{Bidirectional SSM.} We ablate the key bidirectional design of Vim, using ImageNet-1K classification and the Segmenter~\cite{segmenter} semantic segmentation framework on ADE20K. To fully evaluate the power of learned representation on ImageNet, we use a simple Segmenter head with only 2 layers to perform transfer learning on semantic segmentation. We study the following bidirectional strategies. \textit{None}: We directly adopt the Mamba block to process visual sequence with only the forward direction. \textit{Bidirectional Sequence}: During training, we randomly flip the visual sequence. This works like data augmentation. \textit{Bidirectional Block}: We pair the stacked blocks. The first block of each pair processes visual sequence in the forward direction and the second block of each pair processes in the backward direction. \textit{Bidirectional SSM}: We add an extra SSM for each block to process the visual sequence in the backward direction. \textit{Bidirectional SSM + Conv1d}: Based on Bidirectional SSM, we further add a backward Conv1d before the backward SSM (Fig.~\ref{fig:pipeline}).

As shown in Tab.~\ref{tab:aba-dir}, directly adopting the Mamba block achieves good performance in classification. However, the unnatural unidirectional manner poses challenges in downstream dense prediction. Specifically, the preliminary bidirectional strategy of using Bidirectional Block achieves 7 points lower top-1 accuracy on classification. Yet, it outperforms the vanilla unidirectional Mamba block by 1.3 mIoU on semantic segmentation. By adding extra backward SSM and Conv1d, we achieve superior classification accuracy (73.9 top-1 acc \vs~73.2 top-1 acc) and exceptional segmentation superiority (35.9 mIoU \vs~32.3 mIoU). We use the strategy of Bidirectional SSM + Conv1d as the default setting in our Vim block.

\boldparagraph{Classification Design.}  We ablate the classification design of Vim, benchmarking on  ImageNet-1K classification. We study the following classification strategies. \textit{Mean pool}: We adopt mean pooling on the output feature from the last Vim block and perform classification on this pooled feature. \textit{Max pool}: We first adapt the classification head on each token of the visual sequence and then perform max pooling on the sequence to get the classification prediction result. \textit{Head class token}: Following DeiT~\cite{touvron2021training}, we concatenate the class token at the head of the visual sequence and perform classification. \textit{Double class token}: Based on the head class token strategy, we additionally add a class token at the tail of the visual sequence. \textit{Middle class token}: We add a class token at the middle of the visual sequence and then perform classification on the final middle class token.

\begin{table}
\centering
\addtolength{\tabcolsep}{-1pt}
\begin{tabular}{l  | c }
\toprule
Classification strategy  & ImageNet  top-1 acc. \\
\midrule
\textit{Mean pool} & 73.9\\
\textit{Max pool} & 73.4\\
\textit{Head class token} & 75.2 \\
\textit{Double class token} & 74.3\\
\rblue
\textit{Middle class token} & 76.1\\
\bottomrule
\end{tabular}
\caption{Ablation study on the classification design. The default setting for \name{} is marked in \colorbox{blue!10}{blue}.}
\label{tab:aba-cls}
\vspace{-0.15 in}
\end{table}

As shown in Tab.~\ref{tab:aba-cls}, experiments show that the middle class token strategy can fully exploit the recurrent nature of SSM and the central object prior in ImageNet, demonstrating the best top-1 accuracy of 76.1.

\vspace{-0.1 in}

\section{Conclusion and Future Work}
We have proposed Vision Mamba (\name{}) to explore the very recent efficient state space model, \ie, Mamba, as generic vision backbones. Unlike prior state space models for vision tasks which use hybrid architecture or equivalent global 2D convolutional kernel, \name{} learns visual representation in the sequence modeling manner and does not introduce image-specific inductive biases. Thanks to the proposed bidirectional state space modeling, \name{} achieves data-dependent global visual context and enjoys the same modeling power as Transformer, while having lower computation complexity. Benefiting from the hardware-aware designs of Mamba, the inference speed and memory usage of \name{} are significantly better than ViTs when processing high-resolution images. Experiment results on standard computer vision benchmarks have verified the modeling power and high efficiency of \name{}, showing that \name{} has great potential to be the next-generation vision backbone.

In future works, \name{} with the bidirectional SSM modeling with position embeddings is suitable for unsupervised tasks such as mask image modeling pretraining and the similar architecture with Mamba enables multimodal tasks such as CLIP-style pretraining. Based on the pretrained \name{} weights, exploring the usefulness of \name{} for analyzing high-resolution medical images, remote sensing images, and long videos, which can be regarded as downstream tasks, is very straightforward.

\section*{Impact Statement}
We advance the efficiency of the generic vision backbone.
Any societal consequences or impacts that typically relate to work focused on increased efficiency also apply here, as such work necessarily improves the practicality of vision backbone for an array of visual applications with high-resolution input images.

\section*{Acknowledgement}
This work was partially supported by the National Science and Technology Major Project under Grant No. 2023YFF0905400 and National Natural Science Foundation of China (NSFC) under Grant No. 62276108. 

We would like to acknowledge Tianheng Cheng,  Yuxin Fang, Shusheng Yang, Bo Jiang, and Jingfeng Yao for their helpful feedback on the draft.

\nocite{langley00}

\bibliography{vim}
\bibliographystyle{icml2024}

\newpage
\appendix
\onecolumn
\section{Visualization}

\begin{figure*}[ht]
    \centering
    \includegraphics[width=1.0\linewidth]{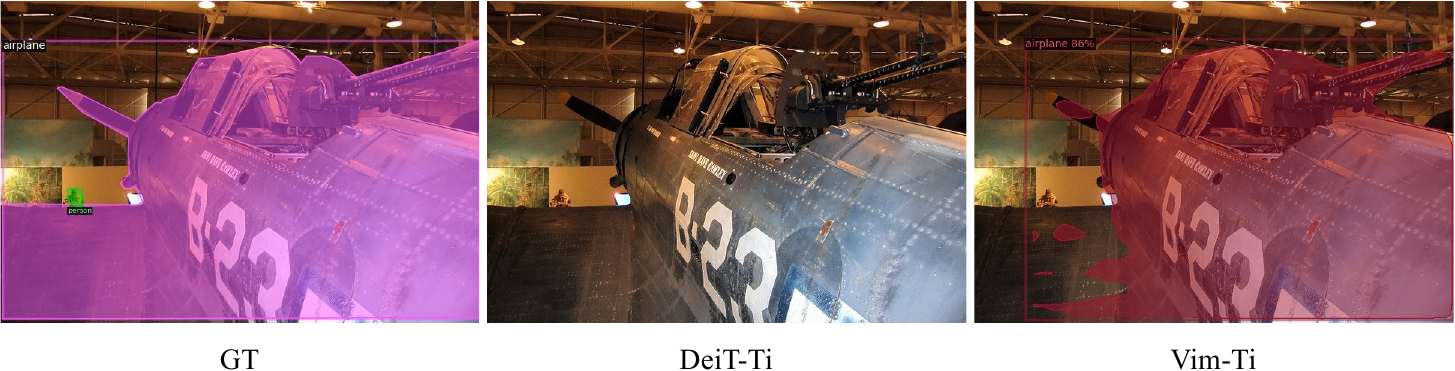}
    \vspace{-0.2 in}
    \caption{Visualization comparison of DeiT-Ti~\cite{touvron2021training} and our Vim-Ti on the Cascade Mask R-CNN~\cite{cai2019cmrcnn} framework. Thanks to the long-range context learning of SSM, we can capture the very large object in the image, which the DeiT-Ti counterpart fails to perceive.
    }
    \label{fig:viscomp}
\vspace{-0.1 in}
\end{figure*}

\section{Additional Setting}
\label{sec: add_setting}

\boldparagraph{Settings for Semantic Segmentation.} We conduct experiments for semantic segmentation on the ADE20K~\cite{zhou2019ade20k} dataset. ADE20K contains 150 fine-grained semantic categories, with 20K, 2K, and 3K images for training, validation, and testing, respectively. We choose UperNet~\cite{upernet} as our base framework. 
In training, we employ AdamW with a weight decay of $0.01$, and a total batch size of $16$ to optimize models. The employed training schedule uses an initial learning rate of $6$$\times$$10^{-5}$, linear learning rate decay, a linear warmup of $1,500$ iterations, and a total training of $160$K iterations. The data augmentations follow common settings, including random horizontal flipping, random re-scaling within the ratio range $[0.5, 2.0]$, and random photometric distortion. During evaluation, we rescale the image to have a shorter side of $512$. 

\boldparagraph{Settings for Object Detection and Instance Segmentation.} 
We conduct experiments for object detection and instance segmentation on the COCO 2017 dataset~\cite{coco}. The COCO 2017 dataset contains 118K images for training, 5K images for validating, and 20K images for testing. We use the canonical Cascade Mask R-CNN~\cite{cai2019cmrcnn} as the base framework. For ViT-based backbones, we apply extra configurations (\eg, interleaved window \& global attention)  to handle the high-resolution images following ViTDet~\cite{vitdet}. For SSM-based \name{}, we directly use it without any modifications. Other training and evaluation settings are just the same. During training,  we employ AdamW with a weight decay of $0.1$, and a total batch size of $64$ to optimize models. The employed training schedule uses an initial learning rate of $1$$\times$$10^{-4}$, linear learning rate decay, and a total training of $380$K iterations. The data augmentations use large-scale jitter data augmentation~\cite{lsj} to 1024$\times$1024 input images. During evaluation, we rescale the image to have a shorter side of 1024.

\section{Extended Comparison on Hierarchical Architecture }
To further compare with hierarchical architectures, we propose another variant Hier-Vim by replacing shifted local window attention in SwinTransformer with the proposed global bidirectional SSM. We detail the configuration in Tab.~\ref{tab:hier_config}

\begin{table}[h!]
    \centering
    \begin{tabular}{l c  c c r r}
    \toprule
      Model & \#Blocks &  \#Channels  & Params  \\ \midrule
      Hier-\name{}-T &  [2, 2, 5, 2] & [96, 192, 384, 768] & 30M \\
      Hier-\name{}-S &  [2, 2, 15, 2] & [96, 192, 384, 768]  & 50M \\
      Hier-\name{}-B &  [2, 2, 15, 2] & [128, 256, 512, 1024] & 89M \\
        \bottomrule
      \end{tabular}%
  \caption{Detailed configurations of different variants of Hier-\name{}. We provide the number of channels and blocks in 4 stages.} 
    \label{tab:hier_config} 
  \end{table}%

\begin{table}[!h]
\centering
\small
\begin{tabular}{l | c  c  | c }
\toprule
Method &  \begin{tabular}[c]{@{}c@{}}image \\ size\end{tabular} & \#param.  & \begin{tabular}[c]{@{}c@{}}ImageNet \\ top-1 acc.\end{tabular} \\
\midrule
Swin-T~\cite{liu2021swin} & $224^{2}$ & 28M  & 81.2\\
FocalTransformer-T~\cite{yang2021focal} & $224^{2}$ & 29M  & 82.2\\
CVT-21~\cite{wu2021cvt} & $224^{2}$& 32M & 82.5\\
MetaFormer-S35~\cite{yu2022metaformer} &$224^{2}$ & 31M  & 81.4\\
GFNet-H-S~\cite{rao2021global}& $224^{2}$ &32M & 81.5\\
\rblue
\textbf{Hier-Vim-T}& $224^{2}$ &\textbf{30M} & \textbf{82.5}\\
\midrule
Swin-S~\cite{liu2021swin} & $224^{2}$ & 50M & 83.2 \\
FocalTransformer-S~\cite{yang2021focal} & $224^{2}$ & 51M&83.5\\
MetaFormer-S35~\cite{yu2022metaformer}&	$224^{2}$ & 73M&	82.5\\
GFNet-H-B~\cite{rao2021global}&	$224^{2}$ &	54M	&82.9\\
\rblue
\textbf{Hier-Vim-S}&	$224^{2}$ &	\textbf{50M}&\textbf{83.4}\\
\midrule
Swin-B~\cite{liu2021swin}&	$224^{2}$ &	88M&83.5\\
FocalTransformer-B~\cite{yang2021focal}&	$224^{2}$ &90M&	83.8\\
\rblue
\textbf{Hier-Vim-B}&	$224^{2}$ &	\textbf{89M}&	\textbf{83.9}\\
\bottomrule
\end{tabular}
\caption{Comparison with hierarchical architectures on ImageNet-1K validation set.}
\label{tab:hier_clscomp}
\vspace{-0.1 in}
\end{table}

\boldparagraph{Classification on ImageNet.} Following the standard training and validation protocols~\cite{liu2021swin,vmamba}, we compare Hier-Vim with popular hierarchical architectures across tiny, small, and base model sizes in Tab.~\ref{tab:hier_clscomp}. The results indicate that Hier-Vim outperforms Swin Transformer by 1.3\% at the tiny size, 0.2\% at the small size, and 0.4\% at the base size, demonstrating competitive performance against well-established and highly-optimized modern hierarchical architectures.

\end{document}